\documentclass[10pt,twocolumn,letterpaper]{article}
\usepackage{iccv}
\usepackage{times}
\usepackage{epsfig}
\usepackage{graphicx}
\usepackage{amsmath}
\usepackage{amssymb}

\usepackage[tableposition=above]{caption}
\usepackage{bbm}
\usepackage{bm}
\usepackage{xspace}
\usepackage[utf8]{inputenc} 
\usepackage{url}            
\usepackage{booktabs}       
\usepackage{amsfonts}       
\usepackage{nicefrac}       
\usepackage{microtype}      
\usepackage{color}
\usepackage[labelformat=empty]{subfig}
\usepackage{algorithm}
\usepackage{algorithmic}
\usepackage{array}
\usepackage{multirow}
\usepackage[para,online,flushleft]{threeparttable}
\usepackage{float}
\usepackage{enumitem}
\usepackage{makecell}
\usepackage{gensymb}         
\usepackage{mathtools}

\newcommand{\bd}{{\mathbf{d}}}

\newcommand{\bo}{{\mathbf{o}}}

\newcommand{\br}{{\mathbf{r}}}

\makeatletter
\DeclareRobustCommand\onedot{\futurelet\@let@token\@onedot}
\def\@onedot{\ifx\@let@token.\else.\null\fi\xspace}

\def\eg{\emph{e.g}\onedot} 
\def\ie{\emph{i.e}\onedot}

\def\etal{\emph{et al}\onedot}
\makeatother

\newcommand{\threesixty}{360$\degree$~}

\usepackage{xcolor}
\usepackage{ifthen}
\usepackage{textcomp}
\definecolor{red}{rgb}{0.9,0.1,0}
\definecolor{slateblue}{rgb}{0.7,0.35,0.9}
\definecolor{green}{rgb}{0, 0.4, 0}
\definecolor{brown}{rgb}{0.3, 0.2, 0}
\definecolor{mahogany}{rgb}{0.75, 0.25, 0.0}
\definecolor{purple}{rgb}{0.7, 0, 0.7}
\definecolor{darkgreen}{rgb}{0, 0.4, 0}
\definecolor{frenchblue}{rgb}{0.0, 0.45, 0.73}
\definecolor{blue}{rgb}{0.0, 0.0, 1.0}
\definecolor{goldenrod}{rgb}{0.65, 0.45, 0.03}
\definecolor{gray}{rgb}{0.5,0.5,0.5}

\newboolean{revising}
\setboolean{revising}{true}
\ifthenelse{\boolean{revising}}
{
    \newcommand{\ignore}[1]{}
    
} {
    \newcommand{\ignore}[1]{}
    
}

\makeatletter
\renewcommand{\paragraph}{%
  \@startsection{paragraph}{4}%
  {\z@}{0.5\baselineskip \@plus 0ex \@minus 0ex}{-1em}%
  {\normalfont\normalsize\bfseries}%
}
\makeatother

\usepackage[pagebackref=true,breaklinks=true,letterpaper=true,colorlinks,bookmarks=false]{hyperref}

\iccvfinalcopy 

\def\iccvPaperID{8314} 

\ificcvfinal\fi

\begin{document}

\title{Moving in a 360 World:\\ Synthesizing Panoramic Parallaxes from a Single Panorama }

\author{
Ching-Yu Hsu  \qquad Cheng Sun \qquad Hwann-Tzong Chen \\

National Tsing Hua University \\
{\tt\small jessie040718@gapp.nthu.edu.tw}
{\tt\small chengsun@gapp.nthu.edu.tw} 
{\tt\small htchen@cs.nthu.edu.tw}

}

\twocolumn[{
\maketitle
\renewcommand\twocolumn[1][]{#1}
  \vspace{-7mm} 
    \centering
    \includegraphics[width=0.95\linewidth]{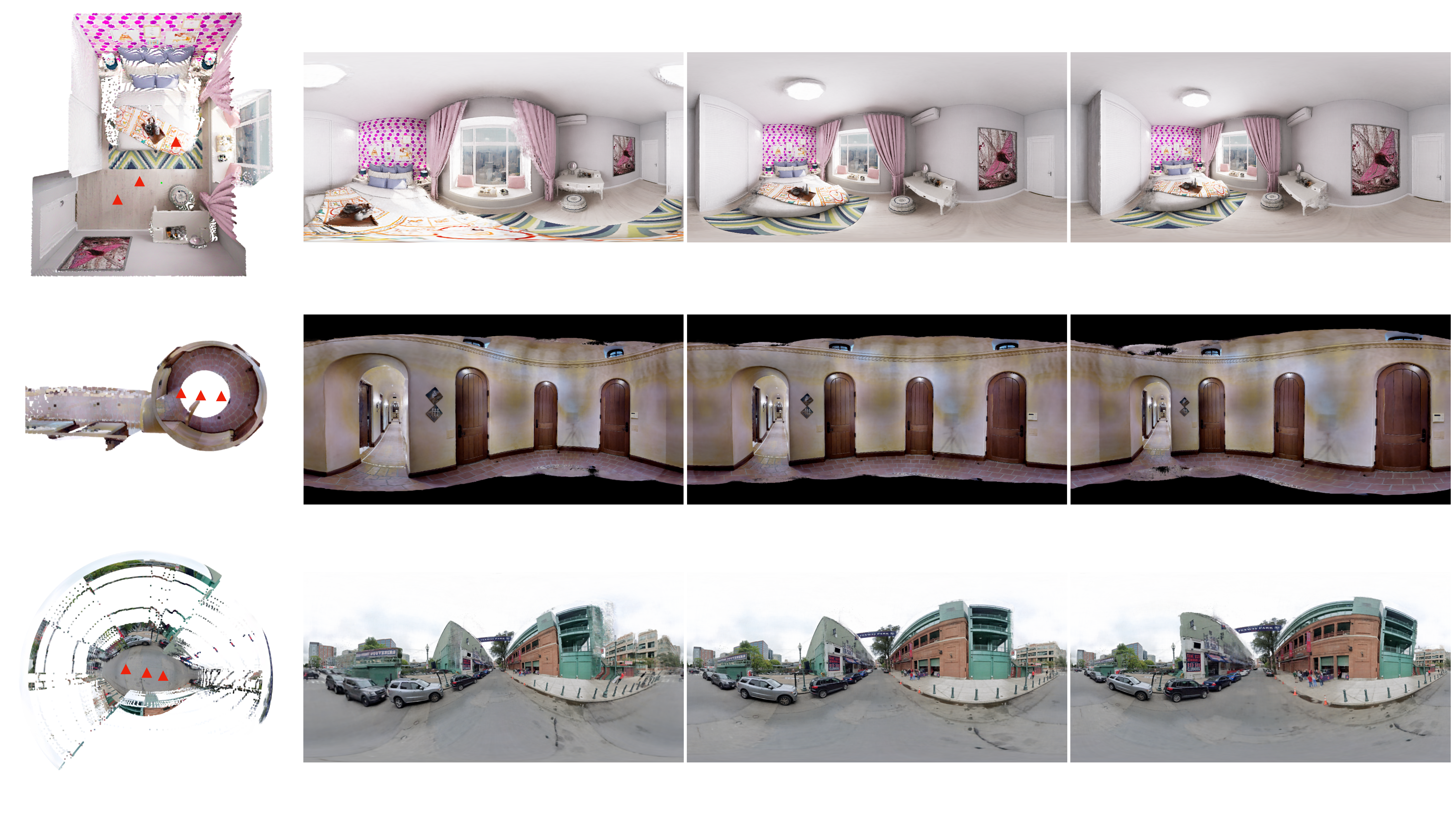}
    \vspace{-2em}
    \captionof{figure}{Synthesizing novel view panoramas from only a single omindirectional input. These panoramas (simulating novel views at the red-triangle spots) are generated with only one \threesixty RGB-D image being given. From top to bottom are scenes from Structured3D, Matterport3D, and Google Street View. Our method is proposed to render panoramic images from arbitrary viewpoints in the scene, and the synthesized results can preserve known color and structure information from the source image.}
  \vspace{7mm} 
}]
\maketitle

\begin{abstract} 
We present Omnidirectional Neural Radiance Fields (OmniNeRF), the first method to the application of parallax-enabled novel panoramic view synthesis. Recent works for novel view synthesis focus on perspective images with limited field-of-view and require sufficient pictures captured in a specific condition. Conversely, OmniNeRF can generate panorama images for unknown viewpoints given a single equirectangular image as training data. To this end, we propose to augment the single RGB-D panorama by projecting back and forth between a 3D world and different 2D panoramic coordinates at different virtual camera positions. By doing so, we are able to optimize an Omnidirectional Neural Radiance Field with visible pixels collecting from omnidirectional viewing angles at a fixed center for the estimation of new viewing angles from varying camera positions. As a result, the proposed OmniNeRF achieves convincing renderings of novel panoramic views that exhibit the parallax effect. We showcase the effectiveness of each of our proposals on both synthetic and real-world datasets.
\end{abstract}
\begin{figure*}[h!]
    \centering
  \includegraphics[width=0.99\linewidth]{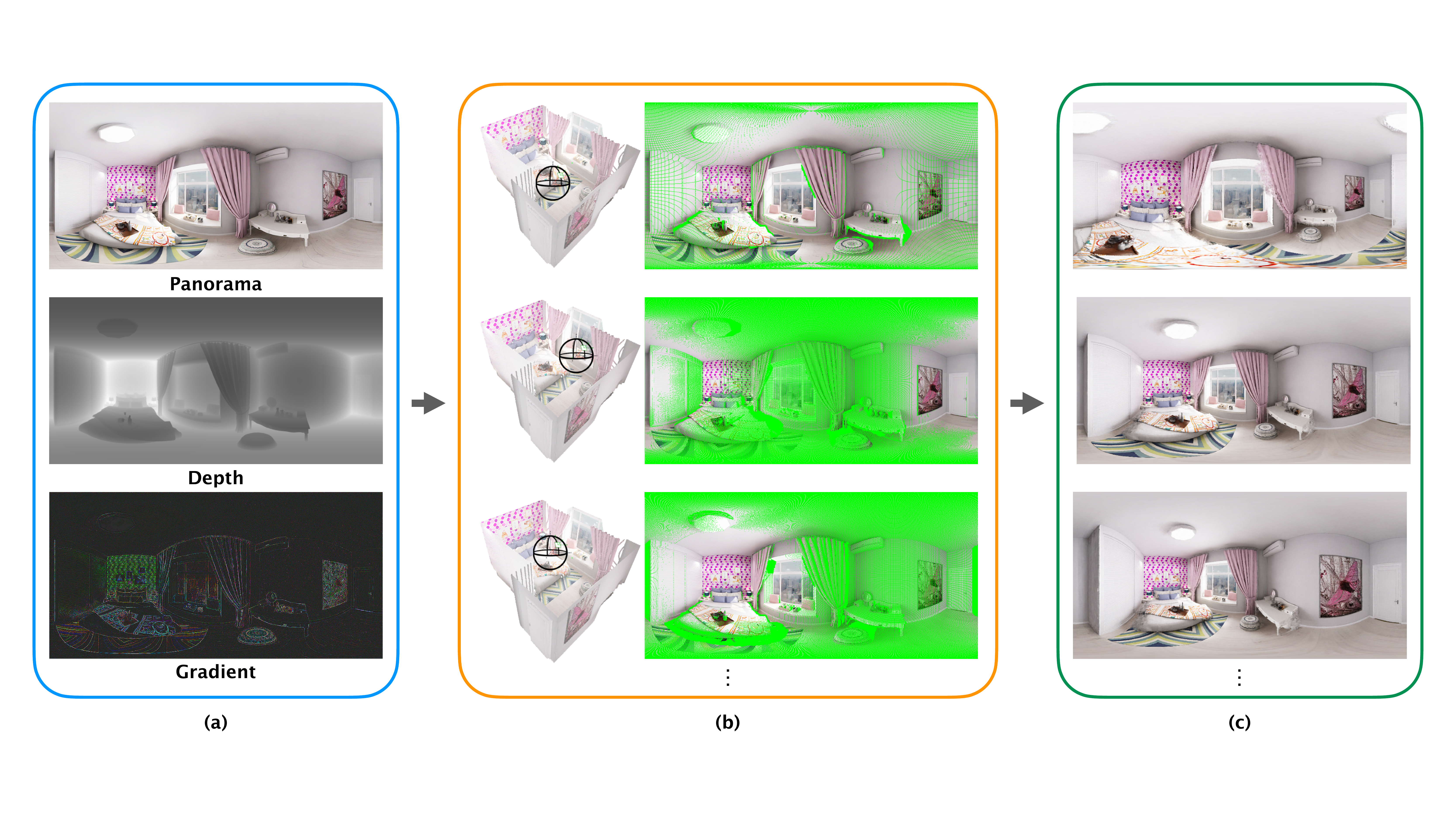}
  \caption{(a) The input data include a panorama, an auxiliary depth map, and a derived gradient image. (b) We generate new training images with various virtual camera poses. Since the information might be missing (marked in green) after re-projection, we could only have partial pixels in each augmented training image. (c) Given new target camera positions, our method is able to generate arbitrary views of the scene.}
  \label{fig:method}
\end{figure*}
\section{Introduction}

Synthesizing novel views with parallax provides immersive 3D experiences \cite{ShumNC05}.
Traditional computer vision solutions employ reconstruction techniques (\eg, structure from motion~\cite{HZ2004} and image-based rendering \cite{0017537,ShumH99}) using a set of densely captured images. However, these approaches suffer from the cost of matching and reconstruction computation for both time and capacity. The recent development in this field focuses on deep learning methods for its strong capability of modeling 3D geometry and rendering new frames. 

While many techniques are proposed to synthesize novel views by taking the perspective image(s) as the input, prior work rarely considers the panorama image as a single source for modeling and rendering. Although perspective images can be acquired conveniently, in order to construct a full scene, it requires a set of dense samples. Furthermore, additional camera variables are essential for estimating relative poses and matching. 
Recently, \threesixty cameras have become more easily accessible, with a growing number of panoramas shared on media and \threesixty datasets released. In a snapshot, it provides an omnidirectional field-of-view, facilitating structure inference and 3D modeling~\cite{ZhangSTX14}. 
This work is in the intersection of novel view synthesis and panoramic imaging, and we propose the first method to animate a single static panoramic photo with motion parallax.

We aim to take the full advantage of a single panorama, which collects a set of viewing directions intersecting at a center and thus suffers no matching problems. The panorama can further be projected to 3D coordinates with auxiliary depth information. In this work, we transform the view synthesis problem into a subsampling task respect to the full scene, without using additional images and camera parameters. With the implicit representation of Neural Radiance Field (NeRF)~\cite{mildenhall2020nerf}, the full scene is formed by a continuous function $F$, which maps 5D coordinates $(x, y, z, \theta, \phi)$ to actual color and density. In NeRF, the whole 3D scene is encoded in weights of a deep fully-connected neural network (a multi-layer perceptron model), which takes the ray origin and ray direction as input. For 2D perspective images, emitted viewing directions are computed from camera parameters. Real-world data might be captured by different cameras with distinct parameters, and to calculate these parameters entails an additional effort.
Accordingly, we propose to derive the NeRF-based representation from the viewpoint of \threesixty panorama. The ray origin is simply the coordinate center. Ray directions are the unit vectors from 3D pixel coordinates to the center, which can be obtained by mapping 2D image coordinates $(x,y)$ to its 3D position along horizontal and vertical axis respectively, based on the auxiliary depth information. These unit vectors are fixed for every panorama, which alleviates the complexity of calculating viewing directions for different input images. 

For a reconstruction task, training with merely a single input sample is apparently not sufficient to create convincing results. Methods trained on perspective images usually require $20$ to $100$ or more samples, depending on the size of the scene and the camera moving distance. We propose a method to augment omnidirectional training data from size one to any desired amount. With an auxiliary depth map being provided, we can retrieve a portion of the actual coordinates of the scene, through multiplying the directional unit vectors by the depth values. To generate a training image at an arbitrary camera position, we translate the center to a target position and project the 3D coordinates back to 2D image space, so that we can produce a `likely-to-be' panorama from a novel viewpoint. However, this panorama is incomplete and the information would be missing due to limited resolution and occlusion. Therefore, cracks and seams appear between pixels when the camera moves. We leverage the pixel-based representation in NeRF and present Omnidirectional Neural Radiance Field (OmniNeRF) to solve the problem. With the new OmniNeRF representation we are free to ignore the missing parts caused by camera translation, and only take account of the valid parts for constructing our training data. Fig.~\ref{fig:method} shows an overview of OmniNeRF.

With the operations under OmniNeRF, camera parameters are not needed, and thus we subside the cost and reduce the error for calculating correspondences between different input images in common perspective settings. The partially available 3D coordinates from a single RGB-D panorama enable free camera movements and allow back-and-forth projections between 3D and 2D spaces for augmenting the training data. We show that the proposed OmniNeRF can render visually plausible results on the new application of novel panoramic view synthesis with parallax effects.

\section{Related work}

\paragraph{Novel view synthesis}
Novel view synthesis has a long history in computer vision for reconstructing or modeling a scene from the acquisition of multi-view 2D pictures of the surroundings. Traditionally, the reference of 3D reconstruction could be a precise 3D model or an approximated representation. Previous methods mainly address 3D reconstruction by applying multi-view stereo and warping strategies for aggregating information among images. Structure from motion \cite{HZ2004} is able to produce sparse point clouds and retrieve camera parameters. Previous methods on novel view synthesis often rely on solving structure from motion, including directly using the point cloud and extracted features \cite{PittalugaKKS19}, or leveraging generated camera parameters \cite{li2020crowdsampling}.
Mesh-based approaches are more common in 3D model reconstruction \cite{BadkiGKS20,abs-1802-05384,KanazawaTEM18,KatoUH18,TatarchenkoDB16,WangZLFLJ18}, and they may be used to create \threesixty view around center object. However, acquiring sufficient information to model an entire scene with detailed 3D meshes cannot be easily done in real world, and thus is less practical for applications. Another way to achieve novel view synthesis without the need of 3D model is to represent the world in different scales of multi-plane images (MPI). MPI provides foreground and background information to solve visibility problems better for synthesizing novel views \cite{li2020crowdsampling, MildenhallSCKRN19, ZhouTFFS18} or adapting it into inpainting-like tasks \cite{ShihSKH20}. These methods have proposed various representations for deriving more realistic results by incorporating deep learning techniques. While many previous approaches take the perspective image(s) as the input to synthesize novel views, very little prior work has considered synthesizing novel panoramas from a single panorama, except recent methods like \cite{LuYLKY19}, which uses concentric mosaics and GAN for stereo panorama conversion, but cannot be easily extended to the application of synthesizing freely-moving novel panoramas as our method.  

\paragraph{Neural 3D representation}
Recent research on mapping 3D spatial location to an implicit representation has shown promising results on encoding the entire scene into weights of a multi-layer perceptron model. In particular, the technique of neural radiance field (NeRF) \cite{mildenhall2020nerf}, presented by Mildenhall \etal, has shown its power of rendering complex objects with both high quality and high resolution.
NeRF describes the world as a continuous function $F$ that maps a 5D coordinates $(x, y, z, \theta, \phi)$ to pixel color and density, where the 5D coordinates include the 3D position of a viewpoint and the horizontal and vertical viewing angles from that view point. Furthermore, NeRF jointly adopts \emph{i}) positional encoding \cite{abs-2006-10739} to embed the input into higher frequency domain and \emph{ii}) alpha composition \cite{MildenhallSCKRN19,PennerZ17,PorterD84,SrinivasanTBRNS19} to simulate the formation of color for a ray, and, as a result, can achieve impressive results. 
Niemeyer \etal~\cite{NiemeyerMOG20} propose Differentiable Volumetric Rendering (DVR), which is also aimed at learning implicit representation of continuous 3D shapes. DVR does not require ground-truth 3D geometry and can learn the implicit shape and texture representations simply from multi-view 2D images. 
 
The aforementioned neural rendering methods are based on 2D perspective images. These methods need to acquire a sufficient number of images from the scene for learning the implicit representations. Some NeRF variants allow more dynamic or less constrained settings of data acquisition. \cite{park2020deformable,pumarola2020dnerf,zhang2020nerf}. Some approaches use online in-the-wild photos to augment the dataset for learning to render identical view but at different time or under different lighting \cite{li2020crowdsampling,abs-2008-02268,MeshryGKHPSM19}. In our work, the input and output data are \threesixty panorama images. From a single RGB-D panorama, our method can augment the training data and learn an MLP as implicit scene representation for synthesizing novel \threesixty panoramas at different locations in the scene.

\begin{figure*}
  \centering
  \includegraphics[width=\linewidth]{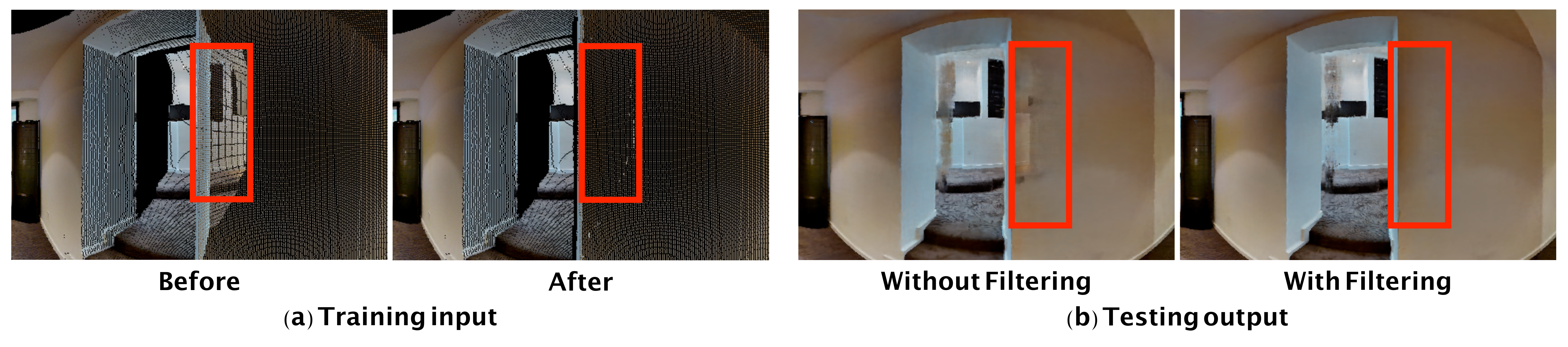}
  \caption{
    \textbf{(a)} An illustration of the process mentioned in Sec.~\ref{ssec:visibility} for filtering out the rays that penetrate the obstacle due to insufficient resolution of the translated view.
    \textbf{(b)} A comparison of the rendered novel view by OmniNeRF trained with and without the filtering out the see-through rays. The rendering quality is greatly improved after solving the visibility issue.
  }
\label{fig:Visibility}
\end{figure*}
\section{OmniNeRF}

We propose OmniNeRF to achieve the goal of synthesizing novel panoramas of a scene at arbitrary viewpoints. The only information available is from an omnidirectional RGB-D image, which would be too limited to train a typical convolutional network directly. The proposed OmniNeRF addresses this issue by augmenting the single RGB-D panorama through back-and-forth projections between 3D world and different 2D panoramic coordinates of different camera poses. The data augmentation mechanism allows us to optimize an Omnidirectional Neural Radiance Field with visible pixels collecting from omnidirectional viewing angles at varying camera locations, and is thus able to train a multi-layer perceptron (MLP) for predicting each pixel in the panorama being viewed from an arbitrary location.

\subsection{Generating training samples}
Generating novel panoramic views with only one omnidirectional image of the scene is a challenging task. Previous perspective-based methods show promising results on scene reconstruction and new viewpoints rendering by using multi-view data with known camera parameters.
Our idea is to adapt a similar process to our scenario by simulating multi-view images from the single RGB-D panorama image. We first produce a set of 3D points from the given RGB-D panorama and then reproject these 3D points into multiple panoramas that correspond to different virtual camera locations.
The generated omnidircetional images are likely to be imperfect as there might be gaps and cracks between pixels due to occlusion or limited resolution. OmniNeRF solves this problem by taking advantage of the pixel-based prediction property of its MLP model, which takes a single pixel rather than an entire image as the input. By considering valid pixels only, we are able to augment our training set given that we have a sparse point cloud of the scene derived from the auxiliary depth map of the input RGB-D panorama.

We project the current panorama into 3D coordinates by the following procedure. First, all pixels can be projected to a uniform sphere by their 2D coordinates. For a pixel $(x, y)$ on the panorama, its vertical and horizontal viewing angles can be defined by $\theta = \pi y / H $, $\phi = 2 \pi x / W $, where $H$ and $W$ are the height and width of the panorama. The   coordinate center would be the current camera position, namely the \emph{ray origin}. Likewise, a \emph{ray direction} simply means a unit vector from the center to the sphere. A novel panoramic view can therefore be determined by moving the camera to a new position and examining what would be sampled on the new sphere by the emitted rays based on the above equations. Not all pixels are supposed to be visible from the new viewpoint. Moreover, the sparse source input might wrongly allow a ray to pass through some occluding object, which causes a visibility problem needed to be solved. (We will come back to address this issue in the next section.) Assuming that we have removed the false visible pixels, we then project the points from the new sphere back to the original pose to obtain the final training image. This transformation is crucial because the scene coordinates are defined by the original source image; the coordinate frame should be kept consistent across all camera poses. As a result, the key of our data augmentation mechanism is to verify which parts of the ground truth will be visible to a given ray origin.

To fully sample the scene, our strategy is to transform views along each axis.
We uniformly sample the camera poses along x-axis and y-axis, within a range of ${[\mathrm{Coord}_\mathrm{min}, \mathrm{Coord}_\mathrm{max}]}$ multiplied a scaling factor $\lambda \in [0,1]$. The range is divided into equal intervals to collect the training views. A smaller $\lambda$ means that samples would be closer to the center. A larger $\lambda$ guarantees better performance while moving farther away from the original center, but the number of applicable pixels would also decrease. We set $\lambda=0.6$ to balance between quality and displacement.
All experiments are trained on $100$ incomplete panoramas derived from a single input panorama and its auxiliary depth map.

\subsection{Visibility} \label{ssec:visibility}
To produce self-simulated multi-viewpoint panoramic images, we transform an image from the original camera pose to any desired pose.
One critical issue is the ambiguity of ray visibility from new viewpoints due to the sparsity of the projected 3D points from only a single image.
More specifically, a ray from a translated camera view could ``see through'' the sparse points of an obstacle and reach a 3D point visible to the original view (see the left-most image in Fig.~\ref{fig:Visibility} for an illustration).
To mask out the ``see-through'' rays, we first apply a median filter on the depth map of the translated view.
Note that only valid pixels on the depth map should be considered.
We then simply filter out pixels whose depth values are larger than the local median depth multiplied by a tolerance ratio (which is set to $1.3$ in this work).
With this simple modification, we can remove most of the incorrect ``see-through'' samples.
A visualization demonstrating the effectiveness of our filtering strategy is shown in Fig.~\ref{fig:Visibility}.

\subsection{Concatenating multiple panoramas}
Although our method only needs one input panorama, it can also combine multiple omnidirectional images once their relative camera positions are known. Since Matterport3D dataset~\cite{ChangDFHNSSZZ17} is the only dataset with multi-view panoramas, we could only apply this experiment on Matterport3D dataset.

\subsection{Regressing with gradient} \label{ssec:reg_depth_grad}
Our initial attempt with the basic setting occasionally suffers from the artifacts of blurry edges, which might come from forcing the model to predict the uncertain regions of the scene. The given training data are not dense enough to cover all areas in the scene; therefore, to render images at new positions would force the model to predict some regions that the model has never seen before. We introduce an additional loss term to improve our model's performance regarding color gradient prediction.
The key of OmniNeRF is to predict unseen pixels between neighboring samples. The model should be able to learn to interpolate from one pixel to another according to ray origin and direction information. 
Inspired by recent depth estimation methods, \eg \cite{LiDCTSLF19}, we include a gradient loss term to enforce the structure-preserving property for color prediction. We use a Laplacian filter to obtain the gradient of the ground truth. The gradient loss can help produce smoother color prediction as well as reduce artifacts. 
However, we are not able to directly compute the output gradient because the input pixels are shuffled (due to 3D reprojection) and thus the neighbor ordering is not maintained. Instead, we add one more head which is parallel to color output in the MLP model to predict gradient. Gradient contains information about neighboring pixels and thus can improve the quality of generated images near boundaries.
\begin{figure*}
     \begin{center}
     \includegraphics[width=\linewidth]{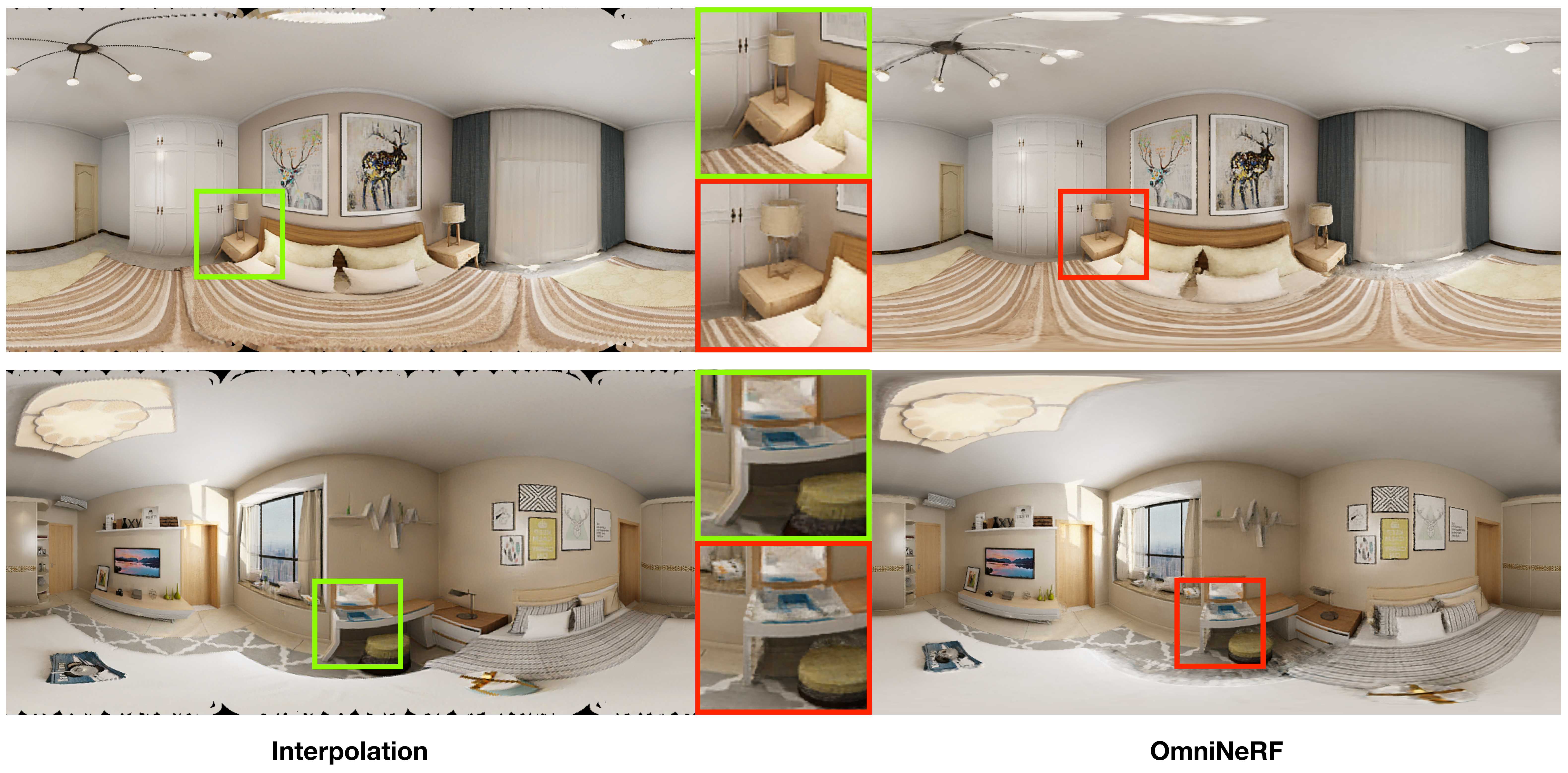}
     \caption{
      Qualitative comparison between interpolation from layout (left) and the proposed OmniNeRF (right).
      The advantages of the interpolation-based method are the rendering speed and the clear appearance with less uncertainty for planar structures.
      However, it is evident that the layout interpolation is unable to render detailed object structures (\eg, the furniture is treated as walls and floors).
      As a result, the generated new views look flat and unnatural, and zigzag patterns can be observed near object boundaries.
      Besides, due to the backward image warping, the generated new view may have some missing pixels.
      In contrast, OmniNeRF generates images with correct structures and minimum artifacts while moving around in the scene.
      It takes a little extra computation cost--- $45$ seconds for a $512 \times 1024$ panorama on one GPU with a batch-size of 1{,}400.
      }
      \label{fig:myLboro}
      \end{center}

\end{figure*}

\subsection{Optimization}
The purpose of an implicit neural representation model is to learn a mapping between 3D coordinates and RGB color space. At each discrete sample on the ray $\br(t) = \bo + t\bd$, where $\bo$ and $\bd$ denote the ray origin and ray direction, the final RGB values $C(\br)$ are optimized from aggregation of color $c_i$ and opacity $\sigma_i$. A positional encoding technique~\cite{abs-2006-10739} is applied to rays for capturing high frequency information. The function of color composition follows the rule in volume rendering~\cite{Max95a}:
\begin{equation}
\begin{aligned}
   \hat{\mathcal{C}}(\br) = \sum_{i=1}^N T_i \left(1-\exp(- \sigma_i \delta_i)\right) c_i,  \\
    \text{where}~  T_i = \exp \left( - \sum_{j=1}^{i-1}\sigma_j \delta_j \right), 
\end{aligned}
\label{equ:color}
\end{equation}
and $\delta_i = t_{i+1} - t_i$ is the interval between two adjacent samples. The overall volume sampling principles are done in a hierarchical way: a `coarse' and a `refined' stage. The coarse and refined networks are identical except the process of sampling pixels on a ray. At the coarse stage, $N_c$ intervals are uniformly sampled alone the ray, while at the refined stage, $N_f$ intervals are decided in accordance with densities from the coarse stage. These two predictions would be optimized by the ground-truth color respectively. 
The overall loss is the sum of two terms: the color loss and the gradient loss. The color loss is simply the $L_2$ loss between the observed color in Eq.~(\ref{equ:color}) and the ground truth. Both the coarse and refined models participate in the optimization process for the composite color. Gradient is aggregated by the same principle as the color term; its loss is also the total squared error between the estimated results and the ground truth. 

\section{Experiments}

\subsection{Datasets}
We test our method on both synthetic and real-world datasets.
In this work, all the panorama images are under equirectangular projection at the resolution of $512 \times 1024$.

\noindent\textbf{Structured3D~\cite{Structured3D} }
Structured3D dataset has 3{,}500 synthetic departments with 18{,}332 photorealistic panoramas rendering.
As the original virtual environment is not publicly accessible, we use the rendered panoramas directly.

\noindent\textbf{Matterport3D~\cite{ChangDFHNSSZZ17} }
Matterport3D dataset is a large-scale indoor real-world 360 dataset, captured by Matterport’s Pro 3D Camera in 90 furnished houses.
The dataset provides 10{,}800 RGB-D panorama images in total, where we find the RGB-D signals near the polar region are missing.

\noindent\textbf{Google Street View~\cite{AnguelovDFFLLOVW10} }
Google Street View images are captured by \threesixty cameras on top of the Street View vehicles or uploaded by users to provide the Google Maps street view services. The accompanied depth is rendered from a facade approximation, so it does not carry detailed information.

For each dataset, we discard scenes that are not feasible for our application, \ie, more than $10\%$ of the pixels do not have depth values.
Since our task is to generate novel panoramic views from a single panorama, consequently, the only supervision to the model comes from those pixels with depths, and it is not applicable to learn from a scene with too many missing depths. 
For Matterport3D, we exclude the polar regions (corresponding to ceiling and floor in most cases) as the RGB-D information is unavailable.

\subsection{Implementation details}
\paragraph{Training protocol}
The Adam optimizer~\cite{KingmaB14} is used for the overall training process.
The learning rate is initialized to $5\cdot 10^{-4}$, which is then exponentially reduced to $5\cdot 10^{-5}$.
The model is trained by 200{,}000 epochs for each experiment with a batch-size of 1{,}400 on a GTX 1080 Ti GPU.
We follow NeRF~\cite{mildenhall2020nerf} to set $N_c=64$ and $N_f=128$ in the coarse and refined networks.
\begin{figure*}
      \centering
     \begin{tabular}{ @{}c@{\hskip 2pt}c@{\hskip 2pt}c@{} }
     \includegraphics[width=.33\linewidth]{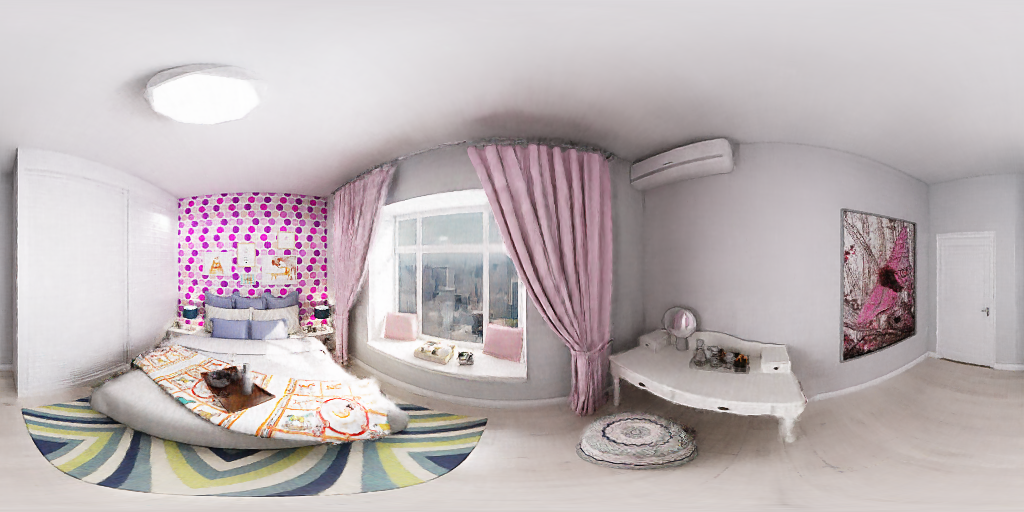}
      &
     \includegraphics[width=.33\linewidth]{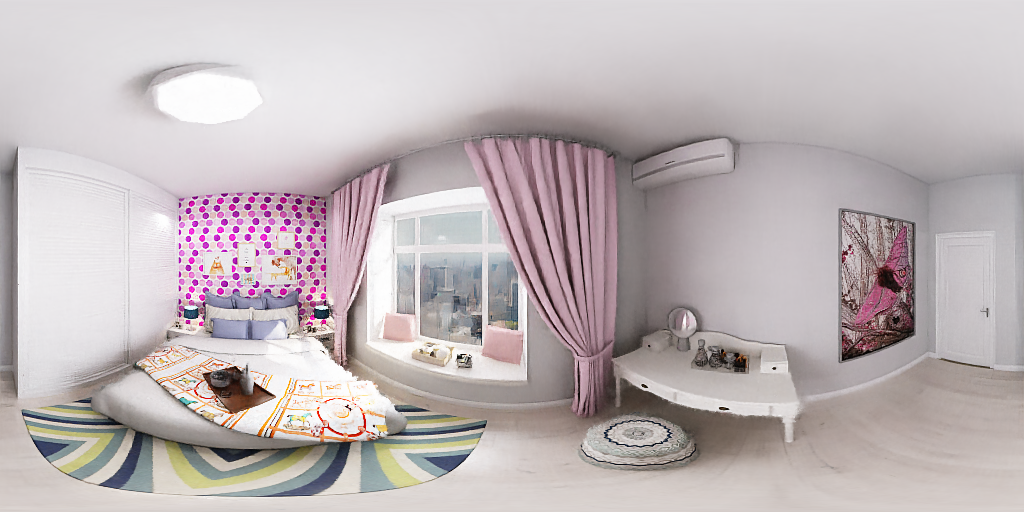}
      & 
     \includegraphics[width=.33\linewidth]{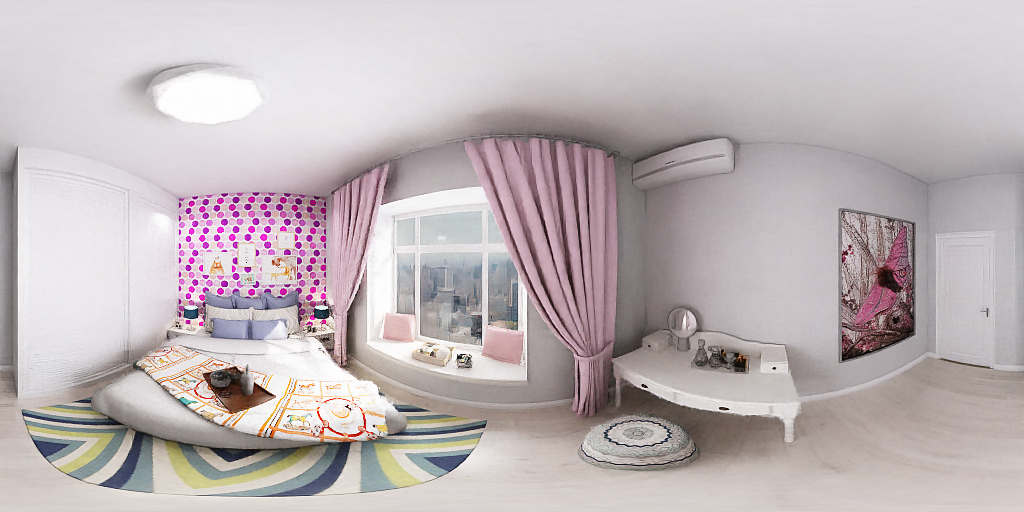}  \\
     \includegraphics[width=.33\linewidth]{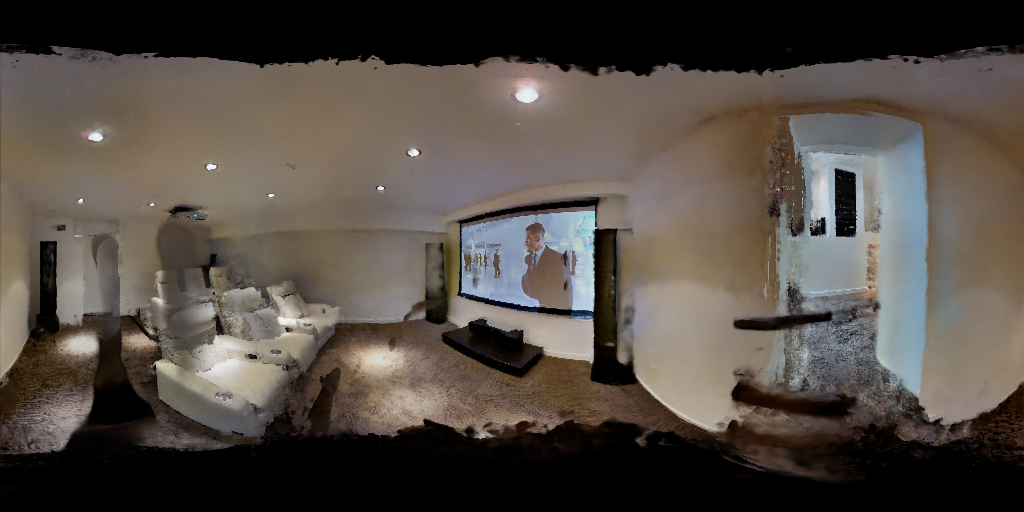}
      &
     \includegraphics[width=.33\linewidth]{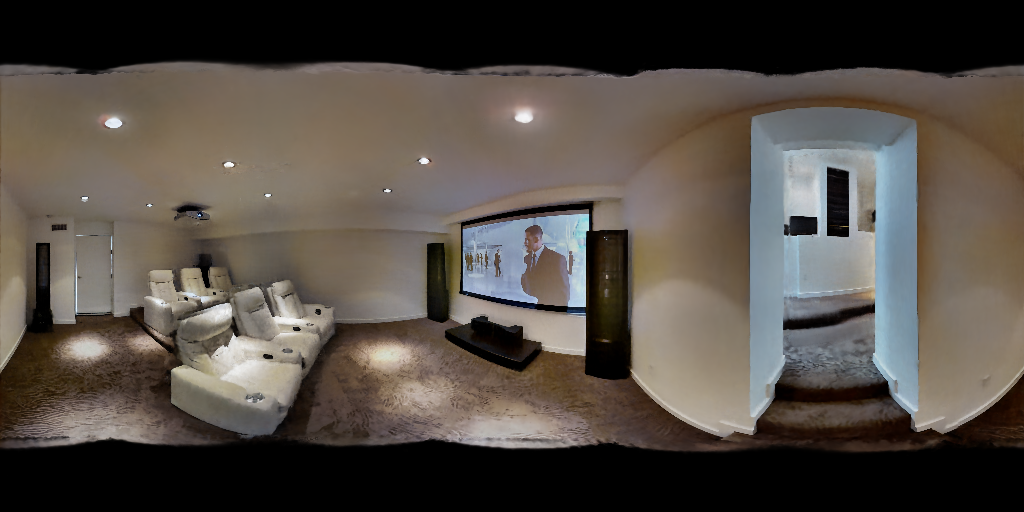}
      & 
     \includegraphics[width=.33\linewidth]{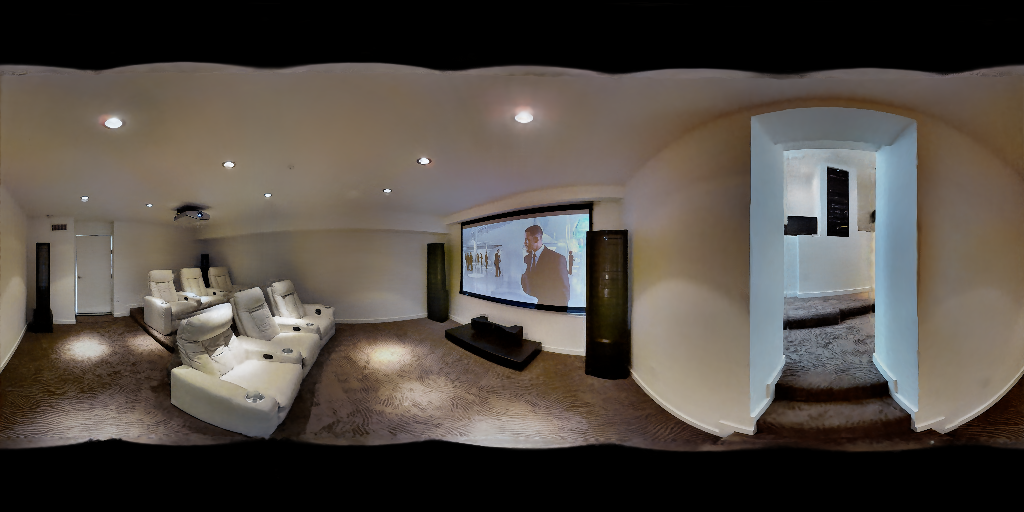}  \\
     \includegraphics[width=.33\linewidth]{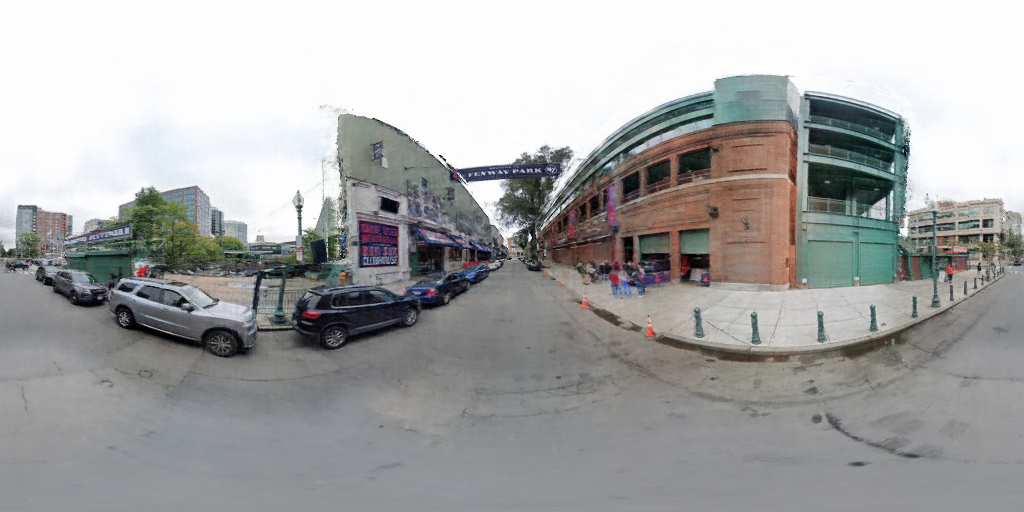}
      &
     \includegraphics[width=.33\linewidth]{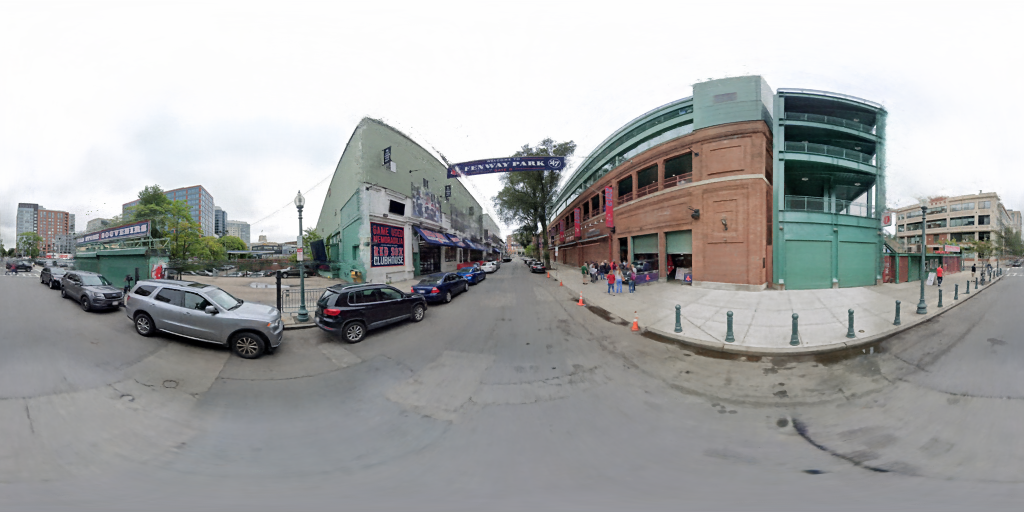}
      & 
     \includegraphics[width=.33\linewidth]{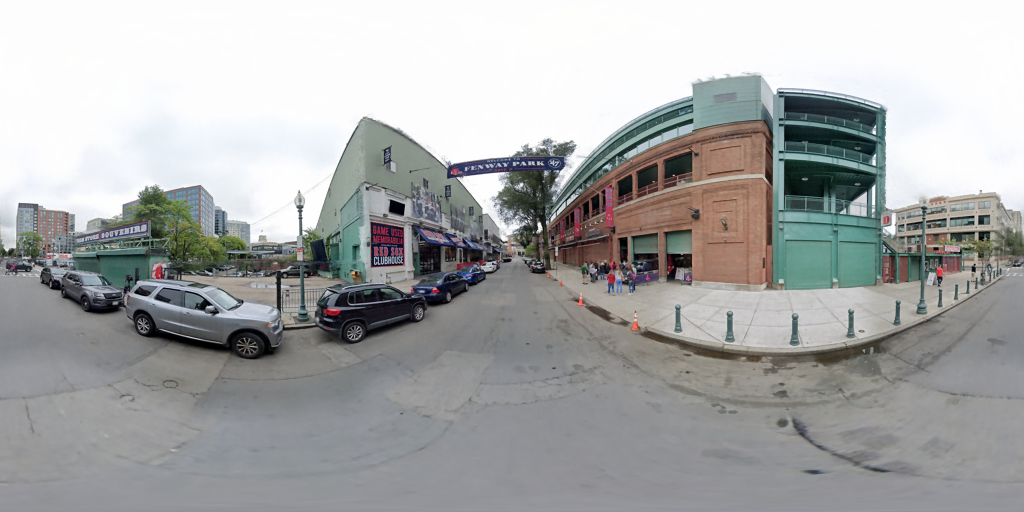}  \\
      (a) NeRF & (b) OmniNeRF (color only) & (c) OmniNeRF (color + gradient)
      \end{tabular}
      \caption{
      The effectiveness of the gradient loss. (a) Outputs trained from a single panorama using NeRF. (b) Outputs optimized from only color loss using OmniNeRF. (c) Rendering results with the gradient loss using OmniNeRF. From top to down are scenes in Structured3D, Matterport3D, and Google Street View. Compared with NeRF, our augmentation method significantly improves the quality throughout all datasets. For synthetic data such as Structured3D, with more accurate depth maps provided, OmniNeRF is able to render images with better quality. For other real-world datasets, like Matterport3D having more complicated indoor structures, with our gradient loss, we are able to reduce artifacts and enhance texture details. Most panoramas in Google Street View are outdoor scenes, and the preciseness of the depth map is not satisfactory. Planer information does not fully accurately correspond with the real structure, for example, some buildings are modeled as background. This situation increases the difficulty of correctly projecting pixels from 3D to 2D image space. Nevertheless, our method is still able to render images closed to the ground truth.}
      \label{fig:ablaf}

\end{figure*}

\paragraph{Evaluation protocol}
The standard evaluation for perspective novel view synthesis is dividing the images of a scene into training and test sets.
Unfortunately, most panoramas we use have no ground-truth nearby panoramic views.
Structured3D rendered panoramas with limited view overlapping.
Some panoramas in Matterport3D do overlap one another. However, the missing pole region in different views highly affect the quality of rendering outputs, so it is not applicable to evaluate quantitative results on adjacent views. 
For Google Street View, the relative camera distances are too far.
For all datasets, we report the statistics by training the model only on translated camera centers and evaluate the generated results at the original camera position with ground truth RGB on account of the lack of ground-truth nearby views.
The quantitative results are reported in Table~\ref{tbl:abla} using PSNR, SSIM~\cite{WangBSS04}, and LPIPS~\cite{ZhangIESW18} to assess the quality of synthesized images.

\subsection{Comparison with baselines} \label{ssec:baseline_construction}
\paragraph{Single image training}
To showcase the proposed learning method's effectiveness, we compare it with NeRF, which is trained on one translated nearby view to serve as the baseline. Based on the quantitative evaluation shown in Table~\ref{tbl:abla}, it is clear that using the original NeRF setting to learn from a single image leads to inferior results.
The proposed OmniNeRF, which learns from a wide range of virtual viewpoints, can significantly outperform the baseline on all datasets and all metrics.

\paragraph{Interpolation from layout}
The layout model captures the gist of a scene's structure, so it is suitable to synthesize new views with camera translation using the layout model.
We first obtain the 3D layout output, which is 2D coordinates for 8 indoor room corners, of each scene by running the HorizonNet~\cite{SunHSC19} pre-trained on the Structured3D dataset; then, given the new camera position, we generate the panorama by interpolation from the 3D layout.
As there are no ground-truth nearby views, we can only compare with the layout interpolation method qualitatively in Fig.~\ref{fig:myLboro}.

\begin{table}
\begin{center}
\begin{tabular}{c@{\hskip 3pt}c||c@{\hskip 3pt}c@{\hskip 4pt}c@{\hskip 4pt}c}
  \hline
  \multirow{2}{*}{\makecell{Learning\\method}} & \multirow{2}{*}{\makecell{Aux. gradient\\loss}} & \multirow{2}{*}{PSNR$\uparrow$} & \multirow{2}{*}{SSIM$\uparrow$} & \multirow{2}{*}{LPIPS$\downarrow$} \\
  &  &  &  &  \\
  \hline\hline
  \multicolumn{5}{l}{\textbf{Structured3D~\cite{Structured3D} dataset}} \\
  \hline
  NeRF & \checkmark  & 22.459 & 0.842 & 0.136 \\ 
  \cline{1-2}
  \multirow{2}{*}{OmniNeRF} & & 33.147 & 0.969 & 0.077 \\
  & \checkmark & 33.249 & 0.968 & 0.073 \\
  \hline\hline
  \multicolumn{5}{l}{\textbf{Matterport3D~\cite{ChangDFHNSSZZ17} dataset}} \\
  \hline
  NeRF & \checkmark & 17.269 & 0.737 & 0.350 \\ 
  \cline{1-2}
  \multirow{2}{*}{OmniNeRF} & & 31.565 & 0.945 & 0.132 \\
  &  \checkmark & 33.943 & 0.965 & 0.108 \\
  \hline\hline
  \multicolumn{5}{l}{\textbf{Google Street View~\cite{AnguelovDFFLLOVW10} dataset}} \\
  \hline
  NeRF & \checkmark & 18.518 & 0.762 & 0.273 \\ 
  \cline{1-2}
  \multirow{2}{*}{OmniNeRF} & & 33.043 & 0.970 & 0.100 \\
  & \checkmark & 33.766 & 0.979 & 0.078 \\
  \hline
\end{tabular}
\end{center}
\caption{
Quantitative comparison on the photorealistic Structured3D dataset, real-world indoor Matterport3D dataset, and outdoor Google Street View dataset.
The ``Gradient'' columns indicate whether the gradient loss introduced in Sec.~\ref{ssec:reg_depth_grad} is employed in the optimization.
The baseline NeRF model learns from the input image as its standard setting, which is detailed in Sec.~\ref{ssec:baseline_construction}.
See Sec.~\ref{ssec:main_exp} for the description of the ablation experiments.}
\label{tbl:abla}
\end{table}

\subsection{Ablation study} \label{ssec:main_exp}
We compare the detailed settings of the objective function for the proposed OmniNeRF in Table~\ref{tbl:abla}.
Our augmented method significantly improve the performance to a plausible level. Learning with additional gradient supervision further uplifts the resemblance between the generated image and the ground truth. Gradient information carries supervision for local structure-preserving, and it also generates more fine details in the scene.
See Fig.~\ref{fig:ablaf} for the visualization.

\begin{figure*}
     \centering
     \begin{tabular}{@{}c@{}}
     \includegraphics[page=1,width=0.99\textwidth]{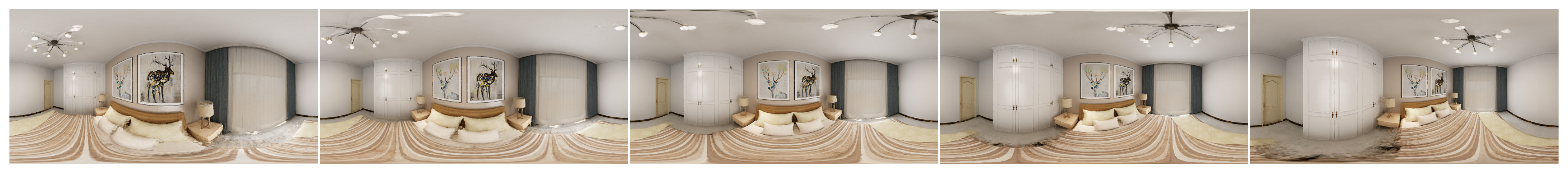} \\ 
     \includegraphics[page=2,width=0.99\textwidth]{images/limitation.pdf} \\ 
     \includegraphics[page=3,width=0.99\textwidth]{images/limitation.pdf} \\ 
     \end{tabular}
     \caption{
     Visualization of OmniNeRF's novel view rendering.
     We show the synthesized novel frames by our method with viewpoints moving along a horizontal path for each scene.
     Three representative examples are shown, where our model is able to render reliable images in most cases.
     See Sec.~\ref{ssec:limitation} for details.
     }
     \label{fig:novelview}
     \vspace{-1.5em}
\end{figure*}

\begin{figure}
    \centering
    \includegraphics[width=.99\linewidth]{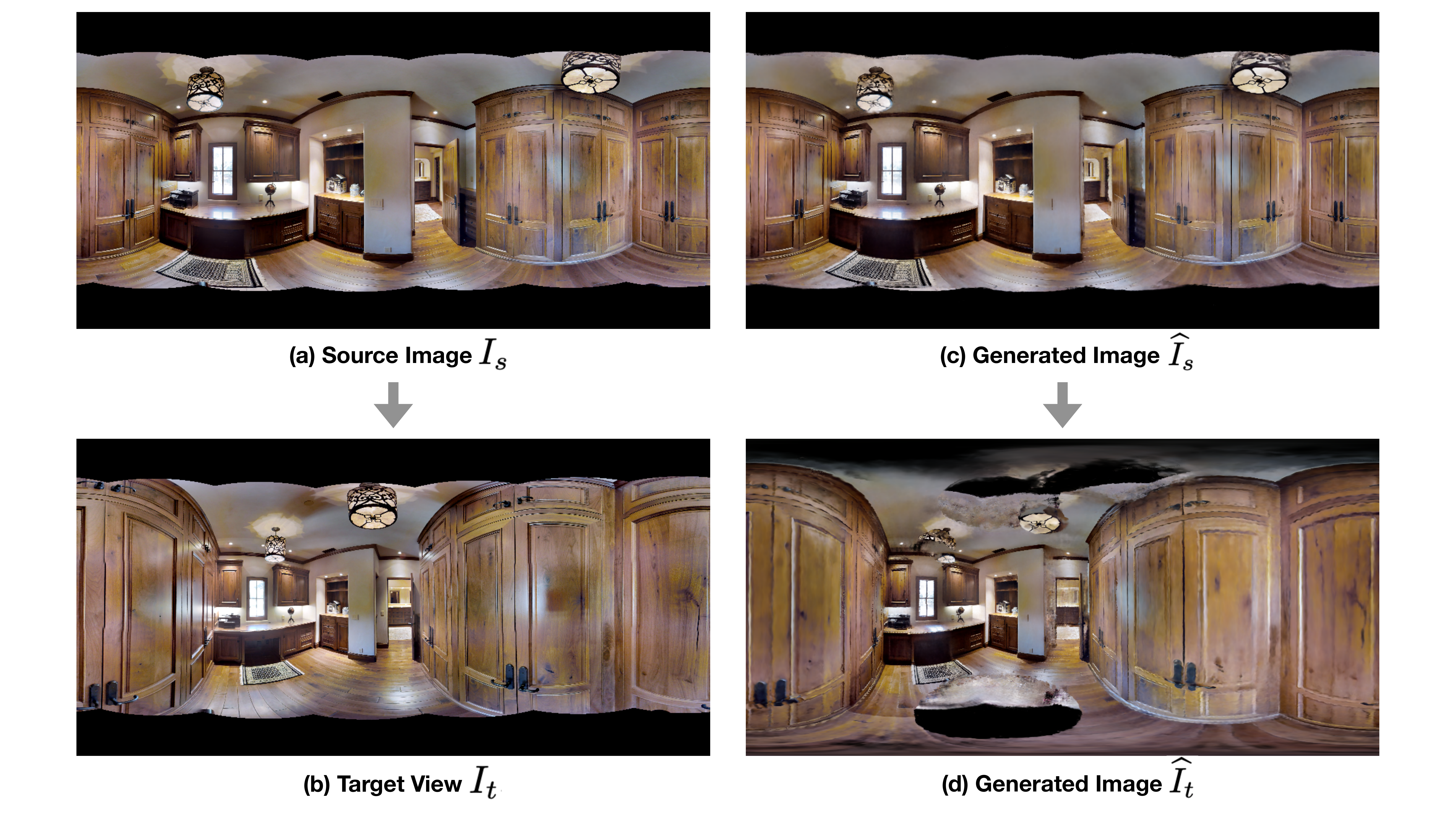}
    \caption{Comparison with real multi-position panoramas in Matterport3D. (a) A source panorama $I_s$ for training. (b) The target view $I_t$, which is the nearest neighbor of $I_s$ in Matterport3D. (c) The OmniNeRF generated panorama $\widehat{I}_s$ that is trained on $I_s$. (d) The OmniNeRF generated panorama $\widehat{I}_t$ by training on $I_s$ and applying the same pose as $I_t$. Even though we are unable to reproduce the precise edges at farther regions, the overall quality is still convincing.
    }
    \label{fig:multiview}
    \vspace{-1.5em}
\end{figure}

\subsection{Comparison with ground truth nearby view}
Some panoramas in Matterport3D~\cite{ChangDFHNSSZZ17} have view overlap.
The distance between cameras can be computed with the provided camera extrinsic.
We select two nearest frame in the same scene to train on one panorama and test on the other.
A qualitative comparison is shown in Fig.~\ref{fig:multiview}.

\subsection{Qualitative results}
We encourage the reader to see the supplementary video demonstration for a fly-through experience provided by our method from just a single panorama.

\subsection{Limitation} \label{ssec:limitation}
To examine the limitation of the proposed OmniNeRF for scene coverage from only a single panorama, we render images at a broader range of camera viewpoints and check the quality of the outcomes.
Specifically, we manually draw a path to render images at some sample positions in a scene. Some representative results on the three datasets with diverse scenes are visualized in Fig.~\ref{fig:novelview}.
For irregular or complicated scenes, the single source image does not contain enough information to infer the occlusion and the opposite side of objects, and thus the results are degraded.
Our model shows better results for simpler scenes with less occlusion, and the range it can render is therefore broader.


\section{Conclusion}
This paper presents OmniNeRF, which learns an implicit representation for \threesixty image rendering from only one single RGB-D panorama. OmniNeRF can synthesize novel panoramic views with the camera moving in the scene. We demonstrate that our augmentation strategy is efficient for providing enough self-simulated multi-viewpoint training samples by leveraging reprojections, transformations, and filtering. Also, it benefits from the property of the MLP model as a pixel-based method, to make use of incomplete appearance. Another advantage of developing rendering techniques for \threesixty images is that no camera parameters are needed in this pipeline, which alleviates the computation cost and the error from matching between different images. The additional gradient loss also improves the performance to generate realistic images.We show that with OmniNeRF, information embedded in a single RGB-D panorama is capable of constructing novel parallax-enabled panoramas.

{\small
\bibliographystyle{ieee_fullname}
\bibliography{egbib}
}

\end{document}